\documentclass[10pt, a4paper]{article}

\usepackage[final]{lrec2026} 
\usepackage{subcaption}
\usepackage{amsmath}
\usepackage{amssymb}
\usepackage{booktabs}
\usepackage{multirow}
\usepackage{xcolor}
\definecolor{midgreen}{RGB}{0,128,0}
\usepackage{enumitem}

\title{SPQ: An Ensemble Technique for Large Language Model Compression}

\name{Jiamin Yao, Eren Gultepe$^*$\thanks{*Corresponding author}}

\address{Dept. of Computer Science, Southern Illinois University Edwardsville, Edwardsville, USA \\
         jamieyao0508@gmail.com, egultep@siue.edu\\}

\abstract{
This study presents an ensemble technique, SPQ (SVD-Pruning-Quantization), for large language model (LLM) compression that combines variance-retained singular value decomposition (SVD), activation-based pruning, and post-training linear quantization. Each component targets a different source of inefficiency: i) pruning removes redundant neurons in MLP layers, ii) SVD reduces attention projections into compact low-rank factors, iii) and 8-bit quantization uniformly compresses all linear layers. At matched compression ratios, SPQ outperforms individual methods (SVD-only, pruning-only, or quantization-only) in perplexity, demonstrating the benefit of combining complementary techniques. Applied to LLaMA-2-7B, SPQ achieves up to 75\% memory reduction while maintaining or improving perplexity (e.g., WikiText-2 5.47 $\rightarrow$ 4.91) and preserving accuracy on downstream benchmarks such as C4, TruthfulQA, and GSM8K. Compared to strong baselines like GPTQ and SparseGPT, SPQ offers competitive perplexity and accuracy while using less memory (6.86 GB vs.\ 7.16 GB for GPTQ). Moreover, SPQ improves inference throughput over GPTQ, achieving up to a 1.9× speedup, which further enhances its practicality for real-world deployment. The effectiveness of SPQ’s robust compression through layer-aware and complementary compression techniques may provide practical deployment of LLMs in memory-constrained environments. Code is available at: \url{https://github.com/JiaminYao/SPQ_LLM_Compression/}
 \\ \newline  }

\begin{document}

\maketitleabstract

\section{Introduction}
Recent advances in large language models (LLMs) have enabled impressive natural language understanding and generation, but their growing size incurs substantial computational and memory costs, making deployment on resource-constrained or real-time systems challenging. Efficient compression techniques that reduce memory while preserving performance are therefore essential.

In this work, we explore an ensemble compression technique that we refer to as SVD-Pruning-Quantization (SPQ), which combines i) variance-based truncation of SVD, ii) activation-based structured pruning, and iii) 8-bit linear quantization. Each component in SPQ is carefully tailored to leverage the strengths of the different compression methods by applying them to where they are most effective. SVD exploits the low-rank structure in attention layers, while pruning removes redundant neurons in the feedforward multilayer perceptron (MLP) layers, thus providing a layer-aware alignment between model structure and compression method. However, SVD and pruning alone degrade perplexity under aggressive compression. To mitigate this effect, we integrate 8-bit linear quantization, which provides an additional compression layer while preserving model performance. By combining these three techniques, SPQ achieves substantial weight memory savings while maintaining or even improving language modeling quality.

Our results demonstrate that careful orchestration of heterogeneous compression methods can yield highly compact models suitable for deployment in memory-constrained environments without significantly degrading performance.

\section{Related Work}
SVD-based compression, structured pruning, and quantization are three commonly used approaches for LLM compression, with each individually offering a distinct advantage, and recently there has been interest in combined methods. 

\subsection{SVD-Based Compression}
Singular Value Decomposition (SVD) has been widely used for neural network compression by approximating large weight matrices with low-rank factors.
Early work showed its effectiveness on convolutional networks \citep{denton2014exploiting}, and later studies confirmed its utility in classical tasks \citep{wang2017research}.
Low-rank factorization also reduces transformer complexity in ALBERT \citep{lan2019albert} and Linformer \citep{wang2020linformer}.
Task-aware methods like Tacoere \citep{guan2024tacoere} exploit cluster structure in transformer weights, while activation-aware SVD (ASVD) \citep{yuan2023asvd} leverages activation statistics for guided truncation.
Adapting SVD to LLMs, SVD-LLM \citep{wang2024svd} and SVD-LLM v2 \citep{wang2025svd} use reconstruction loss-aware decompositions with heterogeneous per-layer ranks, and both apply a LoRA fine-tuning step after decomposition to restore accuracy.
Further refinements include gradient-friendly decompositions \citep{wang2025dobi}, adaptive rank allocation \citep{li2025adasvd}, and applications to Mixture-of-Experts models \citep{li2025structured}.
Overall, these advances highlight SVD’s flexibility, but also its sensitivity to truncation thresholds and layer heterogeneity.

\subsection{Structured Pruning}
Structured pruning removes redundant components in a hardware-friendly manner, such as filters, channels, or attention heads.
Early CNN work ranked filter importance using weight magnitudes \citep{li2016pruning}, pruned neurons based on average activation \citep{hu2016network}, or applied a Taylor-based saliency criterion \citep{molchanov2016pruning}.
For LLMs, SlimLLM \citep{guo2025slimllm} jointly estimates head and channel importance and uses layer-wise sparsity with lightweight regression for quality recovery.
LLM-Pruner \citep{ma2023llm} leverages gradient-based scores and applies LoRA-based tuning with minimal data, while Wang et al. \citep{wang2024pruning} proposed a retraining-free framework that prunes pretrained models before fine-tuning.
As surveyed in \citet{he2023structured}, structured pruning achieves significant FLOP reductions, though some fine-tuning is often needed to maintain accuracy.

\subsection{Quantization}
Quantization reduces numerical precision to compress models and accelerate inference. Early efforts, such as Deep Compression~\citep{han2015deep}, demonstrated the effectiveness of quantization in CNNs, while Q8BERT~\citep{zafrir2019q8bert} extended these ideas to transformers. 
Martynov et al. \citep{martynov2024way} investigate cross-domain properties of transformer weights to design near-lossless quantization schemes, pushing the limits of precision reduction without sacrificing accuracy.
Quantization-aware training (QAT) enables efficient integer-only deployment on edge devices \citep{jacob2018quantization}, while post-training quantization (PTQ) achieves 4–8 bit precision without retraining \citep{banner2019post}. 
For LLMs, GPTQ \citep{frantar2022gptq} performs layer-wise calibration with minimal error and Activation-aware Weight Quantization (AWQ) \citep{lin2024awq} preserves critical activations by using mixed precision and block-wise quantization for stable 4-bit inference.

\subsection{Combined Compression Methods}
Ensemble approaches combine complementary techniques to achieve synergistic compression.
Early work such as Deep Compression \citep{han2015deep} showed that pruning, quantization, and Huffman coding together can dramatically reduce CNN model size without accuracy loss.
Following this idea, several methods extend integration to LLMs: Ren and Zhu \citep{ren2023low} prune redundant heads or neurons before applying low-rank factorization, while SVD-based quantization methods like SVDq \citep{yankun2025svdq} and SVDQuant \citep{li2024svdquant} use low-rank structure to better support mixed-precision or 4-bit quantization. SVD-LLM V2 \citep{wang2025svd} also integrates SVD-based compression with GPTQ quantization to further improve accuracy under tight memory budgets.
QLoRA \citep{dettmers2023qlora} combines 4-bit quantization with low-rank adapters for efficient fine-tuning at near full-precision quality.

\section{Contributions}
Below we highlight how our SPQ framework is different than previous individual or partially-combined compression techniques.

\textbf{SVD in SPQ vs. Prior SVD.}
In contrast to the latest SVD-based method (SVD-LLM v2), which depends on loss approximations for weight matrix types, our SVD variance truncation method is only applied to the attention layers, which were identified experimentally as the most amenable to perplexity retention. This variance-based cutoff yields interpretable, per-layer rank ratios without requiring truncation-loss estimation or group-level allocation.

\textbf{Pruning in SPQ vs. Prior Pruning.}
Prior pruning methods (e.g., SlimLLM, LLM-Pruner) rely on complex importance estimation and multi-structure pruning, increasing implementation cost. SPQ instead prunes only MLP layers, ranks neurons by activation statistics, and derives pruning ratios via log-scale normalization. This yields a lightweight, stable, and hardware-friendly pruning strategy.

\textbf{Quantization in SPQ vs. Prior Quantization.}
Recent quantization approaches like GPTQ and AWQ achieve 4–8 bit precision using layer-wise calibration or activation-aware adjustments, but requires calibration data and solver optimization.
SPQ instead applies post-training 8-bit symmetric linear quantization to all linear layers and supports per-tensor, per-channel, and hybrid scaling without calibration.
This keeps quantization robust, efficient, and fully compatible with SVD and pruning.

\textbf{SPQ Ensemble Technique vs. Prior Combinations.}
Most combined approaches combine only two techniques (e.g., SVDQuant, QLoRA) and omit pruning, which is particularly effective for MLP layers.
Deep Compression applied pruning and quantization to CNNs, but ignored LLM layer heterogeneity.
SPQ is the first to combine SVD on attention layers, activation-based pruning on MLPs, and linear quantization across all layers in a modular, layer-aware pipeline, achieving higher compression with stable perplexity than any single method or two-way combination.

\begin{figure*}[!ht]
\centering
\includegraphics[width=\textwidth]{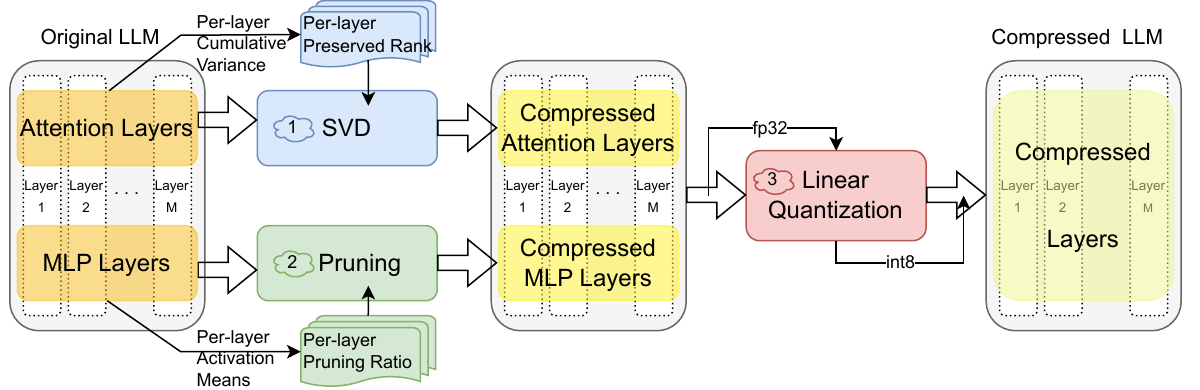}
\caption{SPQ Framework. 1) Attention layers are compressed using variance-retained SVD, where the preserved rank is determined by a variance threshold. 2) MLP layers, per-layer activation means are computed and mapped to pruning ratios via a log-scale formula. 3) Finally, 8-bit linear quantization with different scaling strategies is applied to all linear layers.}
\label{fig:compressed}
\end{figure*}

\section{SPQ Framework}
The following sections provides methodological details regarding our modular compression SPQ framework (Figure \ref{fig:compressed}). Each component is independently configurable and applied to pretrained models without modifying the architecture. SPQ is evaluated on multiple LLM families (OPT, LLaMA, Vicuna, Mistral) ranging from 1B to 7B parameters.

\subsection{Variance-Retained Low-Rank SVD}
We performed low-rank approximation of LLM attention layers using SVD. Let $W \in \mathbb{R}^{m \times n}$ be the weight matrix of a attention layer. The SVD decomposition is
\[
W = U \Sigma V^\top,
\]
where $U \in \mathbb{R}^{m \times m}$ and $V \in \mathbb{R}^{n \times n}$ are orthogonal matrices, and $\Sigma \in \mathbb{R}^{m \times n}$ is diagonal with singular values $\sigma_1 \ge \sigma_2 \ge \cdots \ge \sigma_{\min(m,n)} \ge 0$.

We retain the top-$k_\text{retained}$ singular components that preserve at least an $\epsilon$ fraction of the total variance:
\[
\frac{\sum_{i=1}^{k_\text{retained}} \sigma_i^2}{\sum_{j=1}^{\min(m,n)} \sigma_j^2} \ge \epsilon.
\]
The normalized rank ratio is then
\[
r = \frac{k_\text{retained}}{\min(m,n)},
\]
indicating the compression level. Lower $r$ implies more aggressive compression.

\subsection{Activation-Based Structured Pruning}
We apply structured neuron pruning to the feedforward MLP layers, where entire neurons (rows of the weight matrix and corresponding biases) are removed.

Let $h_j^{(l)}(x) \in \mathbb{R}^{d}$ denote the activation vector of neuron $j$ in layer $l$ for input $x \sim \mathcal{D}$ (calibration dataset). We compute neuron magnitude as
\[
m_j^{(l)} =
\mathbb{E}_{x \sim \mathcal{D}}
\begin{cases}
\frac{1}{d} \sum_{k=1}^{d} |h_{j,k}^{(l)}(x)|, & p = 1,\\[2mm]
\frac{1}{d} \sum_{k=1}^{d} (h_{j,k}^{(l)}(x))^2, & p = 2,
\end{cases}
\]
where $p \in \{1,2\}$: $L_1$ captures mean absolute activation, while $L_2$ emphasizes large activations.

To obtain a per-layer magnitude summary, we compute the mean activation
\[
a^{(l)} = \frac{1}{N_l} \sum_{j=1}^{N_l} m_j^{(l)},
\]
where $N_l$ is the number of neurons in layer $l$.

Layer-wise pruning ratios $r^{(l)}$ are then derived from $a^{(l)}$ using one of three normalization strategies:
\begin{itemize}[noitemsep, topsep=0pt]
    \item \textbf{Linear inverse scaling}
    \[
    n^{(l)} = \frac{a^{(l)} - \min_l a^{(l)}}{\max_l a^{(l)} - \min_l a^{(l)} + \epsilon}
    \]
    \[
    r^{(l)} = r_{\min} + (1-n^{(l)})(r_{\max} - r_{\min})
    \]
    
    \item \textbf{Log-inverse normalization}
    \[
    \ell^{(l)} = \log(a^{(l)} + \delta)
    \]
    \[
    n^{(l)} = \frac{\max_l \ell^{(l)} - \ell^{(l)}}{\max_l \ell^{(l)} - \min_l \ell^{(l)} + \epsilon}
    \]
    \[
    r^{(l)} = r_{\min} + n^{(l)}(r_{\max} - r_{\min})
    \]
    
    \item \textbf{Sigmoid decay}
    \[
    s^{(l)} = \frac{1}{1 + \exp(k(n^{(l)}-0.5))}
    \]
    \[
    r^{(l)} = r_{\min} + s^{(l)}(r_{\max} - r_{\min})
    \]
\end{itemize}
Here, $\epsilon = 10^{-8}$ prevents division by zero, $\delta = 10^{-6}$ avoids $\log(0)$, $k = 10$ controls sigmoid steepness, and $r_{\min}, r_{\max}$ define the pruning ratio range.

\subsection{Post-Training Linear Quantization}
We apply 8-bit symmetric linear quantization to all linear layers. This post-training strategy preserves structural modifications (e.g., pruning and SVD), remains compatible with LoRA fine-tuning, and consistently reduces weight memory by about 25\% while maintaining perplexity.

For a given weight matrix $W \in \mathbb{R}^{m \times n}$, quantization proceeds as:
\[
\begin{aligned}
&\text{scale:} && s_i = \frac{\max(|W_i|)}{2^{b-1}-1}, \\
&\text{quantize:} && \hat{W}_{i,j} = \mathrm{round}\Big(\frac{W_{i,j}}{s_i}\Big), \\
&\text{dequantize:} && W_{i,j}^{\text{quant}} = \hat{W}_{i,j} \cdot s_i,
\end{aligned}
\]
where $b$ denotes the bitwidth ($b=8$ for SPQ).

Per-channel quantization often yields lower perplexity than per-tensor \citep{chitsaz2024exploring}, but incurs higher computational cost. 
To balance accuracy and efficiency, we introduce hybrid quantization, where each layer selects per-tensor or per-channel based on sensitivity. We consider five quantization modes:
\begin{table}[h]
\centering
\small
\setlength{\tabcolsep}{5pt}
\renewcommand{\arraystretch}{1}
\resizebox{\columnwidth}{!}{%
\begin{tabular}{ll}
\toprule
\textbf{Quant Mode} & \textbf{Scale $s_i$} \\
\midrule
Per-tensor   & $s_i$ (one scale per layer) \\
Per-channel  & $s_j$ (one scale per output channel)\\
LNH          & $s_j$ if $l \in \text{Attention}$ else $s_i$ \\
PBH          & $s_j$ if $q^{(l)} \ge P_\alpha$ else $s_i$ \\
MSH          & $s_j$ if $q^{(l)} \ge \mu + k\sigma$ else $s_i$ \\
\bottomrule
\end{tabular}}
\caption{Quantization modes scale selection rules.}
\label{tab:quantization-mode}
\end{table}

Here, $q^{(l)}$ denotes layer weight sensitivity. $P_\alpha$, $\mu$, and $\sigma$ denote the percentile, mean, and standard deviation of $\{q^{(l)}\}$, and $k>0$ sets the MSH threshold.
LNH (Layer-Name Hybrid) uses per-channel for attention layers and per-tensor for MLP layers; 
PBH (Percentile-Based Hybrid) assigns per-channel quantization to the top-$\alpha\%$ most sensitive layers; 
and MSH (Mean$\pm$Std Hybrid) applies per-channel only when $q^{(l)} \ge \mu + k\sigma$, targeting outliers.

\subsection{LoRA Fine-tuning}
Many compression frameworks—such as structured pruning \citep{han2015deep} and low-rank SVD \citep{denton2014exploiting}—include a fine-tuning stage to restore accuracy after modifying weights. Recent SVD-based methods like SVD-LLM \citep{wang2024svd} and SVD-LLM v2 \citep{wang2025svd} further perform LoRA-style fine-tuning on the SVD-decomposed layers to recover performance. In our pipeline, we apply Low-Rank Adaptation (LoRA) \citep{hu2022lora} as the final step: trainable low-rank matrices are injected into selected linear layers while the original weights remain frozen, enabling parameter-efficient adaptation.

Given a weight matrix \( W \in \mathbb{R}^{d_{\text{out}} \times d_{\text{in}}} \), LoRA adds a low-rank update during training:
\[
W x \quad \longrightarrow \quad W x + \frac{\alpha}{r} B A x,
\]
where \( A \in \mathbb{R}^{r \times d_{\text{in}}} \), \( B \in \mathbb{R}^{d_{\text{out}} \times r} \), and \( \alpha \) is a scaling factor. Only \( A \) and \( B \) are updated during fine-tuning; all original weights remain frozen.

\section{Experimental Design}
\subsection{Base Settings}
\textbf{Hardware.}  
Experiments were conducted on two NVIDIA A100-40GB GPUs.

\textbf{Models.}
We primarily benchmark on LLaMA-2-7B and also evaluate models from the LLaMA, OPT, Vicuna, and Mistral families.

\textbf{Evaluation Metrics.}
We compare original and compressed models using memory consumption ($\uparrow$), perplexity ($\downarrow$), and throughput ($\uparrow$), providing a baseline for evaluating compression efficiency in memory and throughput with realistic hardware.

\textbf{Dataset Accuracy.}  
We evaluate the performance of LLaMA-2-7B compressed by the combination of SVD, pruning, and quantization methods across multiple benchmark datasets.

\subsection{SPQ Hyperparameters}

\textbf{SVD.}
For each projection layer, we evaluate multiple variance thresholds $\epsilon \in [0.96, 0.84]$
and compute the retained rank $k_\text{retained}$ as outlined earlier. The normalized rank ratio $r$ is reported per layer.

\textbf{Pruning.}
We sweep the maximum pruning ratio $r_{\max} \in [0.05, 0.30]$, while setting $r_{\min}=0.0$ allowing highly active layers to remain unpruned.
We compute neuron magnitudes using two norms on activations, $L_1$ (mean absolute) and $L_2$ (mean squared), averaged over a small calibration set.
To convert layer-wise activation statistics into pruning ratios, we test three scaling strategies: linear inverse scaling, log-inverse normalization, and sigmoid decay.

\textbf{Linear Quantization.}  
We use 8-bit symmetric linear quantization and evaluate five quantization modes: per-tensor, per-channel, and three hybrid strategies (PBH, LNH, MSH) that selectively apply per-channel quantization to more sensitive layers.

\textbf{Ensemble Techniques.} 
To explore how different compression techniques interact, we apply SVD to attention layers, structured pruning ($L_1$-based and log-scaling) to MLP layers, and linear quantization to all linear layers. We evaluate all combinations with detailed results provided in Section \ref{subsec:pairwise_combinations}. All parameterization is available in our Github.

\section{Experimental Results}
\subsection{SVD Experiments}
\label{subsubsec:svd-experiments}

\begin{figure}[h]
\centering
\begin{minipage}{0.95\linewidth}
    \centering
    \includegraphics[width=\linewidth]{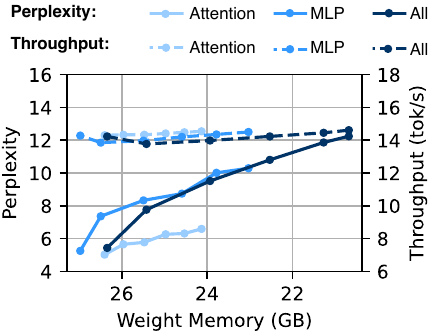}
\end{minipage}
\caption{Per-layer SVD performance.}
\label{fig:svd-layer-results}
\end{figure}

Figure~\ref{fig:svd-layer-results} shows the layer specific application of the variance-based SVD on MLP, attention, and all layers. Variance-based SVD on attention layers have a lower perplexity, as seen by the lowest values, compared to SVD on MLP layers and all layers when compared in the same weight memory size. The throughput also has the best performance on attention layers, which is the highest tokens/sec dashed line compared to the other two dashed lines for the MLP layers and all layers.

\subsection{Pruning Experiments}
\label{subsubsec:pruning-experiments}

\begin{figure}[h]
\centering
\begin{minipage}{0.95\linewidth}
    \centering
    \includegraphics[width=\linewidth]{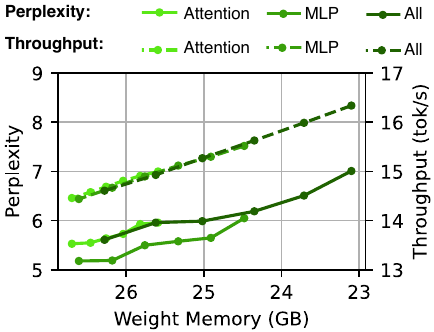}
\end{minipage}
\caption{Per-layer structural pruning performance.}
\label{fig:prune-layer-results}
\end{figure}

In most common structured pruning methods, simply targeting all layers will cause an increase in perplexity. We tested activation-based and layer-scaled pruning on MLP, attention, and all layers with different max pruning ratios using the linear-scaling method, which is shown in Figure~\ref{fig:prune-layer-results}. The resulting perplexity on the MLP layers was lower than the attention and all layers in the same weight memory size. The throughput shows a linear result for different layers, which essentially has the same trend as the weight memory lines. For both perplexity and throughput lines, the MLP layers have a slight advantage if the goal is to balance both perplexity and throughput, especially with larger models.

\begin{figure}[t]
\centering
\begin{minipage}{1\linewidth}
    \centering
    \includegraphics[width=\linewidth]{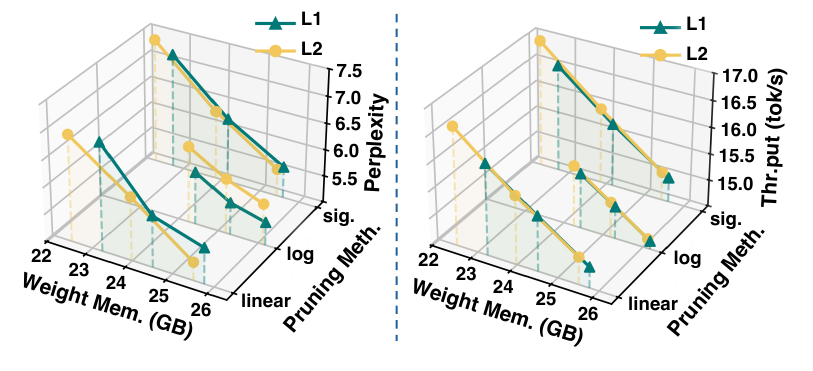}
\end{minipage}
\caption{Pruning performance with different per-layer ratios and methods.}
\label{fig:prune-results}
\end{figure}

Figure~\ref{fig:prune-results} shows that increasing the maximum pruning ratio (i.e., more pruning) raises both perplexity and throughput. $L_1$ and $L_2$ activations produce different per-layer pruning ratios, leading to slight differences in perplexity, but similar throughput and weight memory. Overall, log-scale pruning with $L_1$ offers the best perplexity–efficiency trade-off, and is used in all subsequent experiments involving pruning.

\subsection{Quantization Experiments}

\begin{figure}[h]
    \centering
    \includegraphics[width=0.9\linewidth]{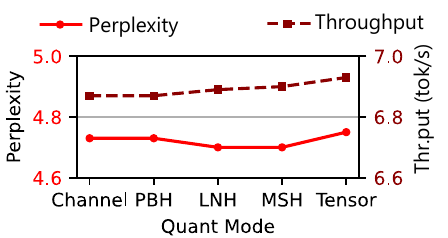}
    \caption{8-bit linear quantization performance.}
    \label{fig:quant-results}
\end{figure}

\begin{figure*}[!ht]
    \centering
    \includegraphics[width=1\linewidth]{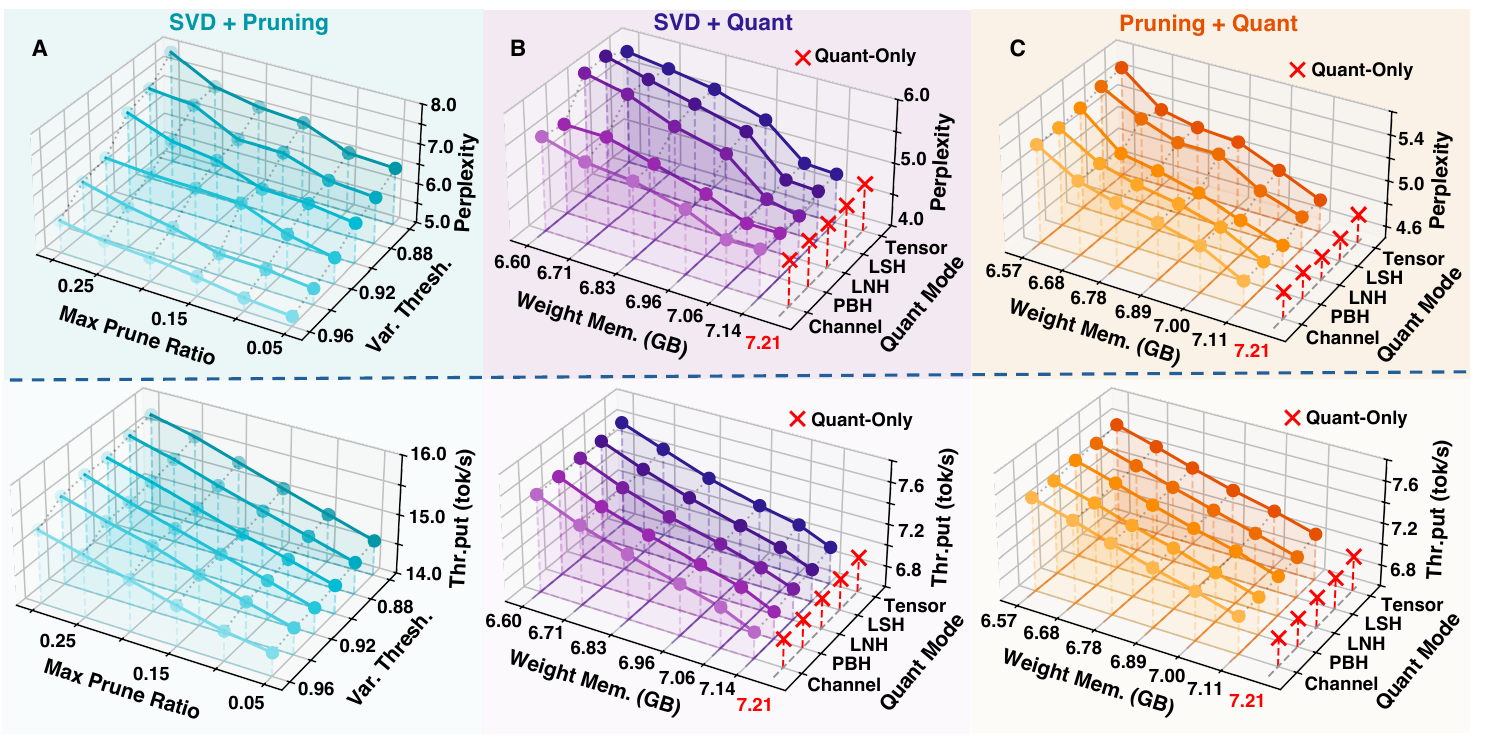}
    \caption{Comparison of ensemble compression methods. 
Top row: Perplexity. 
Bottom row: Throughput. 
}
\label{fig:combined-methods}
\end{figure*}

We apply 8-bit linear quantization, which keeps weight memory fixed at 7.21 GB regardless of per-tensor, per-channel, or hybrid modes (PBH, LNH, MSH) since all modes of quantization store the same number of weights and the extra per-channel modes incurs negligible overhead. Perplexity is nearly identical across all 5 modes. Per-channel quantization slightly improves perplexity at the detriment of lower throughput, as compared to per-tensor, due to finer-grained quantization. The hybrid schemes (PBH, LNH, MSH) mix per-tensor and per-channel quantizations in a balanced manner, providing even lower perplexity than per-channel, while providing a higher throughput. This is due to adapting the type of channel or tensor scaling at each layer. Although quantization reduces throughput relative to full precision, the hybrid settings offer the best trade-off between accuracy and speed.

\subsection{Pairwise Combinations of Compression Methods}
\label{subsec:pairwise_combinations}
The following pairwise combinations: (1) SVD with pruning, (2) SVD with linear quantization, and (3) pruning with linear quantization were performed to reveal how the different methods complement one another and provide insight for our final combined SPQ framework. Figure~\ref{fig:combined-methods} summarizes perplexity and throughput results, while Table~\ref{tab:significant-test} reports statistical tests against the quantization-only baseline.

\textbf{SVD + Pruning.}
From experiments on targeting layers, we found that applying SVD to attention projections and structured pruning to MLPs is most effective. 
As shown in Figure~\ref{fig:combined-methods}, controlling the max pruning ratio ($\leq$ 0.30) and keeping the SVD variance threshold above 0.86 allows perplexity to be maintained at a similar level to the FP32 baseline (5.47). Throughput also improves under these settings, indicating that SVD and pruning are complementary.

\textbf{SVD + Quantization.}
For this setting, weight memory remains constant across quantization modes at the same SVD variance threshold. 
Compared to quantization-only, combining SVD with 8-bit linear quantization achieves additional memory reduction while keeping perplexity and throughput changes minimal. 
Statistical results in Table~\ref{tab:significant-test} confirm that memory savings are significant ($t=$-8.25, $p<$.001), while the perplexity difference (4.76 vs.\ 4.72, $t=$1.82, $p=$.954) is not statistically significant.

\textbf{Pruning + Quantization.}
Similarly, combining pruning with quantization provides further memory reduction compared to quant-only baselines at the same maximum prune ratio. 
The statistical results in Table~\ref{tab:significant-test} show significant memory savings ($t=$-8.45, $p<$.001) with no significant increase in perplexity (4.82 vs.\ 4.72, $t=$2.29, $p=$.980). 
This demonstrates that pruning and quantization can be combined without harming model quality.

\begin{table}[h]
\centering
\small
\setlength{\tabcolsep}{2pt}
\renewcommand{\arraystretch}{1.2}
\resizebox{\columnwidth}{!}{%
\begin{tabular}{l c c}
\toprule
\textbf{Variant} & \textbf{MEM (GB)} & \textbf{Perplexity} \\
\midrule
Q    & 7.21 (--, --) & 4.72 (--, --) \\
S+Q  & 7.10 ($t=$-8.25, $p<$.001) & 4.76 ($t=$1.82, $p=$.954) \\
P+Q  & 7.05 ($t=$-8.45, $p<$.001) & 4.82 ($t=$2.29, $p=$.980) \\
\bottomrule
\end{tabular}}
\caption{Weight Memory(MEM): one-sample one-sided t-test vs. constant Quant-only. Perplexity: two-sample one-sided t-test.}
\label{tab:significant-test}
\end{table}

\textbf{Pairwise Combinations Summary.}
Overall, both SVD+Quant and Pruning+Quant yield significant memory reductions while maintaining perplexity close to the quant-only baseline. 
These results highlight the complementary nature of the techniques and motivate their integration into a unified three-way SPQ pipeline.

\subsection{SVD + Pruning + Quantization Experiments}
\begin{table*}[ht]
\centering
\small
\setlength{\tabcolsep}{2pt}
\renewcommand{\arraystretch}{1.50}
\resizebox{\textwidth}{!}{%
\begin{tabular}{l|l|c c|cccccc|cc|c}
\hline
RATIO (MEM.) & 
METHOD & 
WikiText-2\textbf{\textcolor{midgreen}{$\downarrow$}} & 
C4\textbf{\textcolor{midgreen}{$\downarrow$}} & 
Openb. & 
ARC-e & 
WinoG. & 
HellaS. & 
PIQA & 
Average\textbf{\textcolor{midgreen}{$\uparrow$}} & 
TruthfulQA-1\textbf{\textcolor{midgreen}{$\uparrow$}} & 
TruthfulQA-2\textbf{\textcolor{midgreen}{$\uparrow$}} & 
GSM8K\textbf{\textcolor{midgreen}{$\uparrow$}} \\
\hline
\textcolor{gray}{0\% (26.95 GB)} & 
\textcolor{gray}{Original} & 
\textcolor{gray}{5.47} & 
\textcolor{gray}{6.85} & 
\textcolor{gray}{0.31} & 
\textcolor{gray}{0.75} & 
\textcolor{gray}{0.70} & 
\textcolor{gray}{0.51} & 
\textcolor{gray}{0.78} & 
\textcolor{gray}{0.61} & 
\textcolor{gray}{0.24} & 
\textcolor{gray}{0.39} & 
\textcolor{gray}{0.05} \\
\hline
21\% (21.41 GB) & ASVD      & 
6.54 & 7.66 & 0.32 & 0.70 & 0.66 & 0.50 & 0.76 & 0.59 & 0.25 & 0.41 & 0.03 \\
50\% (13.40 GB)  & SparseGPT & 
7.76 & 8.98 & 0.27 & 0.63 & 0.65 & 0.47 & 0.75 & 0.56 & 0.22 & 0.38 & 0.03 \\
73\% (7.16 GB)  & GPTQ      & 
5.48 & 6.66 & 0.34 & 0.74 & 0.70 & 0.51 & 0.79 & 0.62 & 0.22 & 0.34 & 0.04 \\
\hline
75\% (6.86 GB)  & SPQ       & 
4.91 & 7.11 & 0.30 & 0.72 & 0.68 & 0.51 & 0.77 & 0.60 & 0.24 & 0.38 & 0.05 \\
\hline
\end{tabular}}
\caption{Performance of LLaMA-2-7B compressed by SPQ and baselines on: i) 2 language modeling datasets with perplexity ($\downarrow$), WikiText-2 \citep{thoma2025flar} and C4 \citep{raffel2020exploring}; ii) 5 reasoning datasets with individual and average accuracy ($\uparrow$), OpenBookQA \citep{mihaylov2018can}, ARC \citep{clark2018think}, WinoGrande \citep{sakaguchi2021winogrande}, HellaSwag \citep{zellers2019hellaswag}, PIQA \citep{bisk2020piqa}; 2 reasoning datasets with BLEU ($\uparrow$), TruthfulQA-1 and TruthfulQA-2 \citep{lin2021truthfulqa}; and 1 reasoning dataset with Exact Match Flexible-Extract Accuracy ($\uparrow$) GSM8K \citep{cobbe2021training}.}
\label{tab:dataset_results}
\end{table*}

A comparison of the LLaMA-2-7B model compressed using our proposed SPQ method against common approaches such as ASVD, SparseGPT, and GPTQ is shown in Table~\ref{tab:dataset_results}. ASVD (ratio = 21\%) and SparseGPT (ratio = 50\%) can only achieve low compression without significant performance degradation, whereas SPQ attains substantially higher compression (ratio = 75\%) while preserving or even improving accuracy. SPQ also achieves the lowest weight memory (6.86 GB) among all methods that maintain stable perplexity and task accuracy, reducing weight memory by an additional 2\% compared to GPTQ (7.16 GB), which also uses \texttt{int8} precision. SPQ yields a lower (better) perplexity on WikiText-2, slightly higher (worse) C4 perplexity, and similar accuracy on the reasoning datasets compared to GPTQ. Thus, the overall performance remains comparable to both GPTQ and the original uncompressed model. Moreover, SPQ demonstrates strong results on TruthfulQA and GSM8K. For example, with 54\% more compression than SparseGPT, SPQ is able maintain same BLEU a exact match accuracies as the uncompressed original model. Thus, indicating that SPQ effectively balances memory and model quality using complementary compression techniques to achieve even higher compression than individual advanced SVD, pruning, and quantization strategies.

\begin{figure}[h]
\centering
\begin{minipage}{1.0\linewidth}
    \centering
    \includegraphics[width=\linewidth]{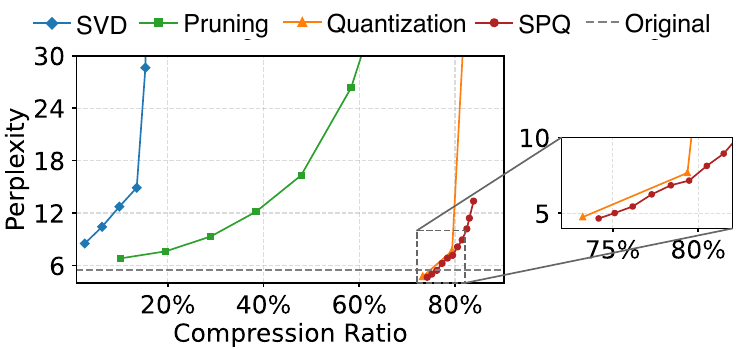}
\end{minipage}
\caption{SPQ and individual method perplexities.}
\label{fig:llama7-ppl-results}
\end{figure}

On the WikiText-2 benchmark with LLaMA-2-7B, we observe distinct behaviors across individual and combined compression methods (Figure~\ref{fig:llama7-ppl-results}). Both SVD and pruning achieve large memory savings, but still lead to sharp perplexity increases at about the 15\% and 40\% compression ratios, respectively, reflecting the limitations of excessive structural sparsity or rank reduction. 
By contrast, linear quantization maintains stable perplexity at 8-bit and 6-bit precision, though performance collapses under 4-bit quantization, which is about 85\% compression ratio. However, SPQ achieves a much better trade-off between compression and performance. The SVD and pruning method added to 8-bit linear quantization can achieve almost the same compression ratio as 4-bit linear quantization, while maintaining a better perplexity. As shown in Figure ~\ref{fig:llama7-ppl-results}, SPQ consistently maintains perplexity below 15 even when the model size is reduced by more than 80\%, demonstrating strong robustness to aggressive compression. The results confirm that combining lightweight pruning and low-rank approximation with quantization helps preserve language modeling quality while substantially reducing memory consumption.

\begin{table}[ht]
\centering
\small
\setlength{\tabcolsep}{2pt}
\renewcommand{\arraystretch}{1.15}
\resizebox{\columnwidth}{!}{%
\begin{tabular}{l c c c c c c}
\toprule
\textbf{Model}
& \multicolumn{3}{c}{\textbf{MEM (GB)}} 
& \multicolumn{3}{c}{\textbf{Perplexity}} \\
\cmidrule(lr){2-4} \cmidrule(lr){5-7}
& Before & After & Ratio & Before & After & Change \\
\midrule
LLaMA-3.2-1B & 5.99 & 2.27 & 62\% & 7.88  & 8.62  & +0.75 \\
LLaMA-3.2-3B & 14.42 & 4.77 & 67\% & 6.86  & 6.39  & -0.47 \\
LLaMA-2-7B & 26.95 & 7.05 & 74\% & 5.47  & 4.71  & -0.76 \\
OPT-1.3B & 5.68 & 1.72 & 70\% & 14.80 & 12.76 & -2.04 \\
OPT-2.7B & 11.12 & 3.11 & 72\% & 12.81 & 11.49 & -1.31 \\
OPT-6.7B & 27.47 & 7.29 & 73\% & 11.02 & 10.04 & -0.98 \\
Vicuna-7B & 26.94 & 7.06 & 74\% & 7.38 & 5.31 & -2.07 \\
Mistral-7B & 28.95 & 7.52 & 74\% & 5.15 & 5.52 & +0.37 \\
\bottomrule
\end{tabular}}
\caption{WikiText-2 perplexity and weight memory of LLaMA, OPT, Vicuna, and Mistral models before and after applying SPQ (MLP pruning with max ratio 0.05, attention SVD with 0.94 variance threshold, and 8-bit linear quantization).}
\label{tab:models_results}
\end{table}

We further evaluate SPQ across models of different sizes and architectures as shown in Table~\ref{tab:models_results} for the LLaMA and OPT model families, as well as Vicuna-7B and Mistral-7B models. The table compares weight memory (MEM) and perplexity before and after applying SPQ. 
Across all models, SPQ substantially reduces memory consumption (62--74\%), while changes in perplexity vary by model. Notably, larger models such as LLaMA-2-7B and OPT-6.7B maintain or even improve perplexity despite aggressive compression (73--74\%), while smaller models show modest increases. These results suggest that SPQ generalizes well across architectures and is especially effective for high-parameter LLMs.

\begin{figure}[h]
\centering
\begin{minipage}{1.0\linewidth}
    \centering
    \includegraphics[width=\linewidth]{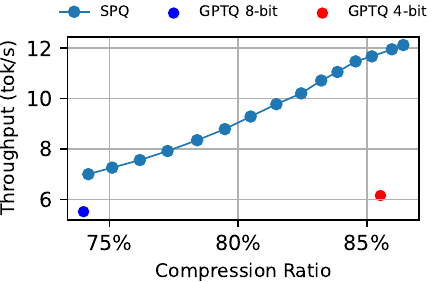}
\end{minipage}
\caption{SPQ and GPTQ throughput comparisons.}
\label{fig:llama7-thr-results}
\end{figure}

The throughput of a compressed model is a critical metric for evaluating its inference efficiency.
In Figure~\ref{fig:llama7-thr-results}, we compare the throughput of GPTQ (8-bit and 4-bit) with our proposed SPQ method on LLaMA-2-7B across various compression ratios.
SPQ consistently outperforms GPTQ: at similar compression levels, it delivers ~1.3× higher throughput than GPTQ-8bit and up to 1.9× higher throughput than GPTQ-4bit.
This demonstrates that SPQ not only reduces memory footprint but also significantly accelerates inference speed, making it more suitable for real-world deployment.
\section{Discussion \& Conclusion}

SPQ is a layer-aware ensemble compression framework that combines SVD low-rank decomposition on attention layers, activation-driven pruning on MLP layers, and 8-bit linear quantization. By assigning each technique to the layer type where it is most effective and determining compression ratios experimentally, SPQ achieves substantial memory reduction while maintaining or even improving perplexity and downstream accuracy. Importantly, SPQ outperforms each individual method (pruning-only, SVD-only, or quantization-only), demonstrating the benefit of combining complementary techniques. Experiments on LLaMA-2-7B shows that SPQ delivers higher compression, faster throughput, and competitive or superior performance compared to strong baselines such as GPTQ.

\textbf{Memory Efficiency:}
Models with larger parameter sizes (e.g., LLaMA-2-7B, OPT-6.7B) required significantly more memory compared to smaller models like LLaMA-3.2-1B and OPT-1.3B. After applying SPQ, all models show a reduction in weight memory, as indicated by the After column for MEM. Notably, LLaMA-3.2-1B experiences the most substantial memory reduction (from 5.99 GB to 2.27 GB), resulting in a weight memory compression ratio of 62\%. Even the larger models such as OPT-6.7B and LLaMA-2-7B show notable memory reductions (73\% and 74\%, respectively).

\textbf{Perplexity Improvement:}
In Table~\ref{tab:models_results}, perplexity indicates a model performance for next-token prediction, LLaMA-3.2-1B has the highest initial perplexity (7.88), which increases slightly to 8.62 after SPQ (+0.75); this small increase is expected due to the limited capacity of such a small model. In contrast, LLaMA-2-7B and OPT-1.3B show negative perplexity changes (-0.76 and -2.04, respectively), indicating that SPQ improves performance while also reducing memory. Interestingly, SPQ’s perplexity gains are more noticeable on larger models, whereas its impact on smaller models like LLaMA-3.2-1B is less pronounced.

\textbf{Throughput Improvement:}
Overall, SPQ delivers consistently higher inference throughput than GPTQ across all compression ratios while preserving model quality. Even when both methods reach similar memory budgets, SPQ yields substantially faster token generation, which provides up to 1.3× speedup over GPTQ-8bit and 1.9× over GPTQ-4bit. This demonstrates that SPQ produces more computation-efficient compressed models than pure quantization, making it a superior choice for real-time and high-throughput deployment.

\textbf{Compression Speed:}
We also compare end-to-end compression time on LLaMA-2-7B at a 75\% compression ratio. Under identical settings, GPTQ requires about 10 minutes, whereas SPQ with 200-step LoRA phase finishes in only 8 minutes. Despite performing additional steps (pruning and SVD), SPQ is 20\% faster due to its simple compression allocation: pruning ratios come from activation statistics and SVD ranks from variance preservation, avoiding the iterative or gradient-based procedures used in prior methods. Thus, SPQ not only delivers higher accuracy and faster inference, but also achieves competitive or superior compression speed compared to GPTQ.

\section{Limitation \& Future Work}

While SPQ demonstrates strong results with its specific combination of SVD, pruning, and quantization, several limitations remain that provide avenues for future research.

\textbf{Ensemble Configuration:} The effectiveness was tied to this particular ensemble design, which may not be universally optimal. Replacing one of the components with alternative techniques (e.g., mixed-precision quantization, structured sparsity, or knowledge distillation) could yield better or worse trade-offs, and only systematic experimentation can determine the best configurations. 

\textbf{Role of Fine-tuning:} Although LoRA is additional processing step among many, it is consistent with existing literature on SVD quantization and model pruning, our framework utilizes it briefly as a final recovery step. It is important to note that even with this 200-step LoRA phase, the end-to-end compression time for SPQ remains lower than that of GPTQ (please see Compression Speed in the Discussion session). This suggests that our approach is not only high-performing but also computationally efficient.

\textbf{Future Work:} Future research will explore more flexible ensemble strategies and evaluate their generalization across diverse model families, downstream tasks, and specialized hardware. Furthermore, we intend to extend our framework to include activation quantization, potentially yielding even greater reductions in memory footprint and operational storage requirements. Another avenue to explore will be the possibility of using other efficient types of matrix factorization techniques.

\section{Bibliographical References}\label{sec:reference}
\bibliographystyle{lrec2026-natbib}
\bibliography{lrec2026-example}

\label{lr:ref}
\bibliographystylelanguageresource{lrec2026-natbib}
\bibliographylanguageresource{languageresource}

\end{document}